\documentclass[journal]{IEEEtran}
\usepackage{cite}

\ifCLASSINFOpdf
\else
\fi

\usepackage{times}
\usepackage{latexsym}
\usepackage{amsmath,amssymb,bbm}
\usepackage{bm}
\usepackage{graphicx}
\usepackage{multirow}
\usepackage{arydshln}
\usepackage{url}
\usepackage{pgfplots}
\usepackage{latexsym}
\usepackage{subfigure}
\usepackage[numbers]{natbib}
\definecolor{blue1}{RGB}{5, 112, 216}
\begin{document}
%
\title{Multi-turn Dialogue \textcolor{black}{Reading} Comprehension with \\ Pivot Turns and Knowledge}
%
%
%

\author{Zhuosheng Zhang, Junlong Li, Hai Zhao
	\thanks{This paper was partially supported by National Key Research and Development Program of China (No. 2017YFB0304100), Key Projects of National Natural Science Foundation of China (U1836222 and 61733011), Huawei-SJTU long term AI project, Cutting-edge Machine reading comprehension and language model (Corresponding author: Hai Zhao).}
	\thanks{Zhuosheng Zhang, Junlong Li, and Hai Zhao are with the Department of Computer Science and Engineering, Shanghai Jiao Tong University, and also with Key Laboratory of Shanghai Education Commission for Intelligent Interaction and Cognitive Engineering, Shanghai Jiao Tong University, and also with MoE Key Lab of Artificial Intelligence, AI Institute, Shanghai Jiao Tong University. (e-mail: zhangzs@sjtu.edu.cn; lockonn@sjtu.edu.cn; zhaohai@cs.sjtu.edu.cn).}
}

%
%

\markboth{IEEE/ACM TRANSACTIONS ON AUDIO, SPEECH, AND LANGUAGE PROCESSING}%
	{}
%



\maketitle

\begin{abstract}
Multi-turn dialogue reading comprehension aims to teach machines to read dialogue contexts and solve tasks such as response selection and answering questions. The major challenges involve noisy history contexts and especial prerequisites of commonsense knowledge that is unseen in the given material. Existing works mainly focus on context and response matching approaches. This work thus makes the first attempt to tackle the above two challenges by extracting substantially important turns as pivot utterances and utilizing external knowledge to enhance the representation of context. We propose a pivot-oriented deep selection model (PoDS) on top of the Transformer-based language models for dialogue comprehension. In detail, our model first picks out the pivot utterances from the conversation history according to the semantic matching with the candidate response or question, if any. Besides, knowledge items related to the dialogue context are extracted from a knowledge graph as external knowledge. Then, the pivot utterances and the external knowledge are combined with a well-designed mechanism for refining predictions. Experimental results on four dialogue comprehension benchmark tasks show that our proposed model achieves great improvements on baselines. A series of empirical comparisons are conducted to show how our selection strategies and the extra knowledge injection influence the results.




\end{abstract}


%
\IEEEpeerreviewmaketitle

\section{Introduction}\label{sec:intro}

\textcolor{black}{Multi-turn dialogue reading comprehension aims to teach machines to read dialogue contexts and solve tasks such as response selection \cite{lowe2015ubuntu,Wu2016Sequential,zhang2018modeling} and answering questions \cite{sun2019dream}, whose common application is building intelligent human-computer interactive systems \cite{Chen2017survey,Shum2018,AliMe,zhu2018lingke}. The task of response selection requires the model to select the appropriate response from a set of candidates given the context of a conversation \cite{lowe2015ubuntu,Wu2016Sequential,zhang2018modeling}. The widely-used benchmark datasets for response selection are the English Ubuntu Dialogue Corpus (Ubuntu) \cite{lowe2015ubuntu} and two Chinese datasets, namely the Douban Conversation Corpus (Douban) \cite{Wu2016Sequential} and E-commerce Dialogue Corpus (ECD) \cite{zhang2018modeling}}. The concerned task has evolved from single-turn matching where only the last utterance in context is used for matching a reply \cite{lowe2015ubuntu,kadlec2015improved,wang2015learning,wan2016match} to multi-turn modeling that benefits from a richer multi-turn context \cite{lowe2015ubuntu,Wu2016Sequential,zhu2018lingke,zhang2018modeling,zhou2018multi,tao2019IOI,gu2019utterance,lin2020grayscale,li58deep,fu2020context}. More recently, the task has further been extended to a \textcolor{black}{multi-choice dialogue machine reading comprehension (MRC)} style setting \cite{Rajpurkar2016SQuAD,trischler2017newsqa} with extra questions---given a dialogue context and corresponding questions, the machine is required to select the appropriate answer accordingly \cite{sun2019dream,cui2020mutual}. 

\textcolor{black}{The solution architectures for both of the dialogue response selection and dialogue comprehension tasks are similar, and existing models generally consist of two components: encoder and matching network.} \textcolor{black}{Most previous works focused on the matching of the context and response \cite{Wu2016Sequential,zhou2018multi,gu2019interactive} where matching signals in each utterance--response pair are fetched from their interaction according to their representations and then aggregated as a matching score.} Benefit from recent advance of pre-trained language models (PrLMs), a critical feature of current models \cite{zhu2020dual,jin2019mmm} is that they choose advanced pre-trained language models (PrLMs) such as BERT \cite{devlin2018bert} and GPT \cite{radford2018improving} as encoder implementation.     

Although the dialogue comprehension shares a similar form with the standard reading comprehension tasks  \cite{zhang2020mrc,Rajpurkar2016SQuAD,trischler2017newsqa,Zhang2018subword,zhang2020retro}, understanding multi-turn dialogues is much more complex for two key challenges. On the one hand, multi-turn dialogues are multi-party, multi-topic, and always have lots of turns in real-world cases (e.g., chat history in social media) \cite{zhang2018modeling,gopalakrishnan2019topical,xu2020topic}. For the scenario of multi-turn dialogue with over 20 turns of utterances, we argue that not all the utterances contribute to the final response selection. The noise in context utterances seriously hurt the matching performance \cite{zhang2018modeling,yuan-etal-2019-multi}. Therefore, matching the response with all the utterances would be suboptimal. On the other hand, people rarely state obvious commonsense explicitly in their dialogues \cite{forbes2017verb}, therefore only considering superficial contexts may ignore implicit meaning and cause misunderstanding.


	\begin{table}
		\centering
		\makeatletter\def\@captype{table}\makeatother\caption{Multi-turn dialogues have different topics (in bold).}
		{
			\begin{tabular}{p{7.2cm}}
				\hline 
				\textbf{Dialogue 1}\\
				\hline
				\textit{W: \textbf{Well, I'm afraid my cooking isn't to your}}  \textit{\textbf{taste.}}                \\
				\textit{M: \textbf{Actually, I like it very much.}}                         \\
				\textit{W: \underline{I'm glad you say that. Let me serve you some more fish.}}
				
				\textit{M: \textbf{Thanks. I didn't know you are so good at cooking.}} \\ 
				
				\textit{W: Why not bring your wife next time?}                                    \\
				\textit{M: OK, I will. She will be very glad to see } \textit{you, too.}                                \\
				\hline
				Question: \textit{What does the man think of the}
				\textit{woman's cooking?}                            \\
				\hline
				\textit{A. It's really terrible.}                                                   \\
				\textit{B. It's very good indeed. *}                                                                  \\
				\textit{C. It's better than what he does.}                                                            \\
				\hline
			\end{tabular}%
		}
		
		\label{tb-example-topic}
	\end{table}

	\begin{table}
		\centering
    	\makeatletter\def\@captype{table}\makeatother\caption{Commonsense is required in the question.}
		{
			\begin{tabular}{p{7.2cm}}
				\hline
				\textbf{Dialogue 2}\\
				\hline
				\textit{\textbf{M: Look at the girl on the bike!}}\\
				\textit{F: Oh, yes she's really a smart girl.}\\
				\hline
				{Question: \textit{Where are the two persons?}}                            \\
				\hline
				\textit{A. At home.}                                                   \\
				\textit{B. In their classroom. }                                                                  \\
				\textit{C. On the street. *}                                                            \\
				\hline
			\end{tabular}
		}
		\label{tb:example-commonsense}
	\end{table}

We demonstrate two examples from DREAM dataset \cite{sun2019dream} in Tables \ref{tb-example-topic} and \ref{tb:example-commonsense}. Table \ref{tb-example-topic} shows the topic shifts in turns, and only a few of them, called pivot turns, are directly related to the question. Treating all turns equally hurts the understanding of multi-turn dialogues as shown in some previous works \cite{zhang2018modeling,yuan2019multi}. Table \ref{tb:example-commonsense} shows that commonsense knowledge is required to answer the question (e.g., bikes are always on the streets). However, such required knowledge usually cannot be obtained or inferred from the given material (passage, question, or answer options) of the task.

To refine the context with topic clues, our prior work \cite{zhang2018modeling} took the last utterance in the context as the pivot to mine the connections with the rest preceding utterances.\footnote{In this work, we define \textit{pivot} as the intermediate utterance(s) used for refining context as the topic or evidence clues.} 
However, the indicative utterance is not always the last one. Besides, there would be more than one pivot utterance in a conversation and there is a lack of focus on commonsense knowledge. In this work, we propose a pivot-oriented deep selection model (PoDS) for dialogue comprehension. In detail, our model first picks out the pivot utterances from the conversation history according to the semantic matching with the candidate response or question, if any. Besides, knowledge items related to the dialogue context are extracted from a knowledge graph as external knowledge. Then, the pivot utterances and the external knowledge are combined together with a well-designed mechanism for giving predictions. Experimental results on four dialogue comprehension benchmark tasks show that our proposed model achieves great improvements on baselines. Our contributions are three folds:

1) We propose a flexible and explainable pivot-oriented deep selection model to locate the informative utterance(s) as the pivot clues to refine the context with topic clues and apply a fine-grained matching with the candidate response.

2) For more advanced dialogue comprehension that requires commonsense reasoning, we introduce external knowledge from a knowledge graph to enrich its representation. 

3) Experimental results on four benchmark corpora show that the PoDS achieves substantial improvements over the baselines. A series of empirical comparisons are conducted to show how our selection strategies and the extra knowledge injection influence the results.


\section{Related Work}\label{sec:related}
\subsection{Pre-trained Language Model}
Recently, deep contextualized language models (PrLMs) have been shown to be effective in learning universal language representations, achieving state-of-the-art results in a series of flagship natural language processing tasks. Prominent examples are Embedding from Language Models (ELMo) \cite{Peters2018ELMO}, Generative Pre-trained Transformer (OpenAI GPT)  \cite{radford2018improving}, BERT \cite{devlin2018bert}, Generalized Autoregressive Pre-training (XLNet) \cite{yang2019xlnet}, Robustly Optimized BERT Pretraining approach (Roberta) \cite{liu2019roberta}, and ALBERT \cite{lan2019albert}. Providing fine-grained contextualized embedding, these pre-trained models can be either easily applied to downstream models as the encoder or used for fine-tuning.

Despite their impressive success, these PrLMs remain limited in representing the contextualized information in the domain-specific corpus because they are usually trained on a general corpus \cite{xu2019bert,whang2019domain,zhang2020SemBERT,zhang2019sg}. Besides, multi-turn conversation modeling is more challenging, which requires deep interaction between the abundant context and response. The simple linear layer used in the PrLMs for downstream task prediction would not be sufficient enough. We need a better strategy of taking advantage of both sides of the pairwise modeling in PrLMs and more effective interaction to infer the relationship between the candidate response and the conversation history. 
One solution is to conduct post-training on task-specific datasets. Another is task-adapted model design, which is the major focus of this work. In the present paper, we extend the deep contextualized PrLMs into multi-turn dialogue modeling with a well-crafted model design.


\subsection{Multi-turn Dialogue Modeling}
Developing a dialogue system means training machines to converse with a human using natural language. Towards this end, a number of data-driven dialogue systems have been designed \cite{lowe2015ubuntu,Wu2016Sequential,zhang2018modeling}, and they can be categorized into two types of architectures: one concatenates all context utterances \cite{lowe2015ubuntu,lowe2017training}
and the other separates and then aggregates utterances \cite{Wu2016Sequential,zhou2018multi,tao2019IOI}. 
Existing works showed that a matching structure is beneficial for improving the connections between sequences in neural network models
\cite{lowe2015ubuntu,lowe2017training,zhou2016multi,wang2015learning,li2020memory,dcmn20}.
Recent work has extended attention to modeling multi-turn response selection. DUA \cite{zhang2018modeling} employed self-matching attention to route the vital information in each utterance. DAM \cite{zhou2018multi} proposed a method based entirely on attention achieving impressive improvement. IOI \cite{tao2019IOI} improved context-to-response matching by stacking multiple interaction blocks.
State-of-the-art methods show that capturing and leveraging matched information at different granularities across context and response are crucial to multi-turn response selection \cite{gu2019interactive,tao2019multi}. 

In contrast with the above studies, the present work benefits from a deep pre-trained Transformer architecture with deep context interaction to model the relationships between context and response. Our proposed PoDS has three differences from models in the studies mentioned above. (1) The PoDS adopts a deep Transformer encoder as the backbone with dual pairwise modeling between the context and response, while the models of previous studies are based on RNN, and the inputs are encoded separately. (2) The PoDS adopts a \textit{pack and separate} strategy to take advantage of both deep Transformer encoders and further response-aware interaction. (3) The PoDS employs more carefully selective matching between the context utterances and response when calculating attention weights in the interactive matching module. A series of empirical comparisons are conducted to show the influence factors.

\subsection{Knowledge Enhanced Language Representation}
	
	Recently, researchers pay more and more attention to enhancing text models with Knowledge Graphs (KGs), since KGs obtain a great amount of systematic knowledge.  Integrating background knowledge in a neural model was first proposed in the neural-checklist model by \citet{kiddon2016globally} for text generation of recipes. \citet{liu2019k} combined knowledge triples in KGs with original texts before modeling them with BERT to get more hidden information. \citet{mihaylov2018knowledgeable} attended to relevant external knowledge and combined this knowledge with the context representation in Cloze-style reading comprehension. \citet{bosselut2019comet} employed the triples in KGs as corpus to train GPT \cite{radford2018improving} for commonsense learning. \citet{lin2019kagnet} proposed a knowledge-aware graph network based on GCN and LSTM with a path-based attention mechanism. \citet{zhang2019ernie} fused entity information with BERT to enhance language representation, which can take advantage of lexical, syntactic, and knowledge information simultaneously.
	
	Some previous works have already taken either key turns \cite{yuan2019multi} or commonsense knowledge \cite{sun2019dream} into account when dealing with multi-turn dialogues. However, none of them make a combination of these two important factors. In this work, we make the first attempt to utilize key turns as pivots and commonsense knowledge simultaneously to enhance the language representation of multi-turn dialogues. 
	Pivot utterances are selected from the conversation history according to the semantic matching with the candidate response, whose representations are directly extracted from the encoded text of the original dialogue context.
	On the other hand, knowledge items related to the dialogue context are extracted from a knowledge graph as external knowledge, which are then encoded with the same PrLM encoder.
	As a result, representations of pivot utterances, commonsense knowledge, and the original text of multi-turn dialogues are in the same vector space so that they can be easily fused together.

\begin{figure*}
	\centering
	\includegraphics[width=1.0\textwidth]{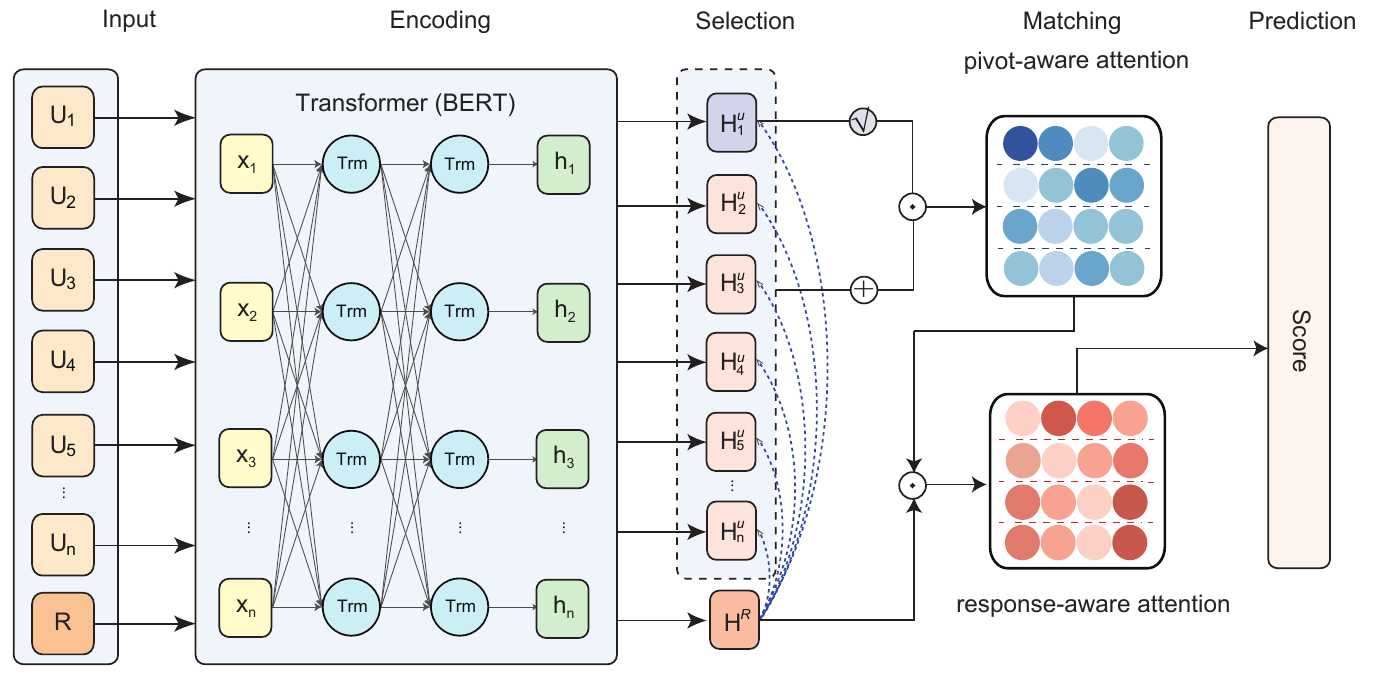}
	\caption{\textcolor{black}{Structural overview of the context—response scoring model for response selection task.}}\label{fig:overview}
\end{figure*}

\section{Pivot-oriented Deep Selection Model}\label{sec:dsm}
In the model section, we first introduce our backbone framework for the widely used dialogue response task, such as the Ubuntu \cite{lowe2015ubuntu}, Douban \cite{Wu2016Sequential}, and ECD \cite{zhang2018modeling} task, whose form is to select the proper response from a candidate list, given the context history composed from various dialogue utterances. Then, we extend the framework to recent conversational reading comprehension, whose major difference is having extra questions, thus the aim is to select the correct answer from various candidate options, given the dialogue context and the corresponding questions, as examples shown in Tables \ref{tb-example-topic}-\ref{tb:example-commonsense}.

Figure \ref{fig:overview} shows our PoDS framework for conventional multi-turn conversation tasks, \textcolor{black}{supposing that the symbol in dark-marked $\textbf{H}^{u}_{1}$ is the selected pivot}. The PoDS formalizes the context and response into a joint input to feed the Transformer encoder \cite{NIPS2017_7181} and mines key information from the utterances and response across the encoding. The PoDS then separates the context and response, semantically matches each utterance, and response to select the pivot utterances. The pivot utterances are used to refine the original conversation context to obtain the pivot-aware contextual representation. The matching module then calculates the attentive interaction between the refined context and candidate response. In the last module, the response-aware contextual vectors are delivered to a Gated Recurrent Unit (GRU) \cite{Cho2014Learning} in chronological order of the utterances in the context, and the last hidden state is passed to a linear layer to obtain the final matching score. 

We denote the training set as a triple  $\mathcal{D}= \left \{\left ( \textbf{C},\textbf{R},\textbf{Y}\right )_i \right \}_{i=1}^{N}$, where $\textbf{C} = \left \{ \textbf{U}_{1},\dots,\textbf{U}_{n} \right \}$ is a conversation context with $\left \{\textbf{U}_{k} \right \}_{k=1}^{n}$ as the utterances. 
$\textbf{R}$ is a response candidate while $\textbf{Y} \in \left \{ 0,1 \right \}$ is a binary label, indicating whether  $\textbf{R}$ is a proper response for $\textbf{C}$.

\subsection{Encoding}
To make the best use of the Transformer-based deep encoders, we employ a \textit{pack and separate} method by first packing the context and response as a joint input to feed the encoder and then separate them according to the positions for further interaction.\footnote{We found this strategy could better take advantage of the benefits from the pairwise interaction of the inputs in BERT. Detailed discussion is shown in Section \ref{match_method}.}

Given the context $\textbf{C}$ and response $\textbf{R}$, tokens are packed into a sequence:
\begin{equation*}
    \textbf{X} = \{\texttt{[CLS]}  \textbf{R}  \texttt{[SEP]}  \textbf{U}_1 \texttt{[SEP]} \dots \texttt{[SEP]} \textbf{U}_n \texttt{[SEP]}\},
\end{equation*}
where \texttt{[CLS]} and \texttt{[SEP]} are special tokens. We separate $\textbf{C}$ and $\textbf{R}$ with \texttt{[SEP]} to guide the model to learn the relationship between the context and response. $\textbf{X}$ is then fed into the BERT encoder, which is a deep multi-layer bidirectional Transformer, to obtain a contextualized representation $\textbf{H}$. 

\subsection{Separation}
To obtain the representation of each individual utterance and response, we split the last-layer hidden state $\textbf{H}$ into $\textbf{H}^R$ and $\textbf{H}^C = \{\textbf{H}^{u}_{1}, \dots, \textbf{H}^{u}_{n}\}$ as the representations of the  response and context, according to its position information. All utterances $\textbf{H}^{u}_{i}$ in the same context are padded to the maximum length $l$ among them.  

\subsection{Selection}\label{method:selection}

To select the pivot utterances from the context, this module scores each utterance with respect to the response.\footnote{It is possible to calculate the similarity with other tokens, such as the special token, \texttt{[CLS]}, which is supposed to carry the global information of the whole sequence, or with the last utterance. Here we only take the response, for example. Detailed discussion is shown in Section \ref{tab:selection_opt}.} The $m$ top-scoring utterances are selected as the topic clues. 

Since both utterance or response are ended with a special token \texttt{[SEP]}, which is supposed to learn the sentence structure after BERT's pre-training through the next sentence objective  \cite{cohan2019pretrained}, we pick it out as the representation of the corresponding utterance or response. Let $\textbf{H}^{u}_{i} (i \in [1,n])$ and $\textbf{H}^R$ denote the utterance representation and response representation, respectively. We calculate the distance between each utterance-response pair to obtain the best related utterance(s): 
\begin{equation}
\textbf{S}_i=\textup{Dist}(\textbf{H}^R, \textbf{H}^{u}_{i}),
\end{equation}
where $\textup{Dist}(\cdot,\cdot)$ is the distance measurement. In this work, we use cosine similarity.


After scoring each utterance, the $m$ top-scoring sentences are selected and concatenated together following the original order to form a pivot context \textcolor{black}{$\textbf{H}^P \in \mathbb{R}^{q \times d} $}, where the context length $q=m\times l$ and $d$ denotes the dimension.

\subsection{Matching}\label{matching}
The matching layer is used to model the relation between the context and response, which contains two parts, 1) we first compute the \textit{pivot-aware attention} to obtain the refined context; 2) we then calculate the \textit{response-aware attention} to estimate the matching relationship between the refined context and response.\footnote{\textcolor{black}{To make fair comparisons, i.e., the analysis in Section \ref{sec:num_pivots}, our baseline also adopts the matching mechanism by taking all the utterances as the pivots.}}

Multi-head attention \cite{vaswani2017attention} is used in this work to capture the relationship between two sequences. We denote it as MHA(·), which is implemented as follows:
\begin{equation}
\begin{aligned}
&\textrm{Att}(E'_Q,E'_K,E'_V)=\textrm{softmax}(\frac{E'_Q(E'_K)^T}{\sqrt{d_{head}}})E'^V,\\
&\textrm{head$_i$}=\textrm{Att}(E_QW^Q_i,E_KW^K_i,E_VW^V_i),\\
&\textrm{MHA}(E_Q,E_K,E_V)=\textrm{Concat(head$_i$...head$_h$)},\\
\end{aligned}
\end{equation}
where $W^Q_i \in \mathbb{R}^{d_{model} \times d_{ head}}, W^K_i \in \mathbb{R}^{d_{model} \times d_{head}}, W^V_i \in \mathbb{R}^{d_{model} \times d_{head}}, E_Q \in \mathbb{R}^{d_q \times d_{model}}, E_K \in \mathbb{R}^{d_k \times d_{model}}, E_V \in \mathbb{R}^{d_v \times d_{model}}$, $d_q,d_k,d_v$ and $d_{head}$ denote the dimension of Query vectors, Key vectors, Value vectors and each head, respectively. $h$ denotes the number of heads. We always assume $d_k=d_v$ and $d_{model}=h \times d_{head}$.
	
In detail, the refined context is produced by taking the pivot presentation $\textbf{H}^P$ as the attention to the context representation $\textbf{H}^{C}$:
\begin{equation}
\begin{split}
\textbf{H}^{CP} = \textrm{MHA}(\textbf{H}^{C}, \textbf{H}^P, \textbf{H}^P),
\end{split}
\label{eq2:Image_Representation}
\end{equation}
where $\textbf{H}^{CP}$ is the weighted sum of all the hidden states and it represents how the vectors in $\textbf{H}^{C}$ can be aligned to each hidden state in $\textbf{H}^P$.

Similarly, we calculate the response-aware attention by taking the response representation $\textbf{H}^R$ as the attention to the refined context $\textbf{H}^{CP}$. Thus, we have \textcolor{black}{$\textbf{H}^G = \textrm{MHA}(\textbf{H}^{CP}, \textbf{H}^R, \textbf{H}^R)$} as the response-aware contextual representation.

\textcolor{black}{In accumulating the response-aware contextual representation $\textbf{H}^G \in \mathbb{R}^{u \times d}$ for the final prediction where  $u = n\times l$, we employ a GRU to propagate information in $\textbf{H}^G$ where each element in the dimension of $u$ is seen as the time step in GRU. Supposing $\mathbf{\hat{H}} = [\hat{h}_1, \dots, \hat{h}_u]$ denotes the hidden states of the input sequence, we have}

\begin{equation}
\mathbf{\hat{H}} = \textbf{GRU}(\textbf{H}^G).
\end{equation}

The last hidden state $\hat{h}_u \in \mathbb{R}^{d}$ is selected for final prediction.


Note that the matching procedure can be regarded as two stages of interaction between the utterances and response. In detail, the joint pairwise input of the context and response interacts with the deep Transformer encoder where the coarse matching information can be fetched. After selecting the pivot utterances from the context, we conduct more fine-grained two-step matching between the response and these utterances for further enhancement.

\subsection{Prediction}
We concatenate the last hidden state $\hat{h}_u$ with the first hidden state $h_0$ of the BERT encoder and feed the result into the output softmax layer to compute the final matching score.\footnote{\textcolor{black}{$h_0$ is regarded as the pooled representation as the BERT output \cite{devlin2018bert}}.} We define $g( \hat{h}_u,h_0)$ as
\begin{equation}
g(\hat{h}_u,h_0) = \textup{SoftMax}(\textbf{W}_g[\hat{h}_u;h_0] + \textbf{b}_g),
\end{equation}
where $\textbf{W}_{g}$ and $\textbf{b}_{g}$ are trainable parameters. During the training phase, model parameters are updated according to a cross-entropy loss. 

\section{Incorporation of Extra Knowledge}\label{model:knowledge}
Now, we extend the above pivot selection framework to the recent advanced conversational reading comprehension task that can benefit from extra knowledge injection. Besides the context $C$ for the concerned multi-turn dialogue MRC represented as $[\textbf{U}_1,\dots,\textbf{U}_{n_u}]$, the model is required to answer the related questions $\textbf{Q}$, by selecting the response from an answer set $\textbf{A} = [\textbf{A}_1,\dots,\textbf{A}_{n_a}]$. In this work, we treat the question and the answer option as an integral \textcolor{black}{through concatenation}, so that we have \textcolor{black}{$\textbf{QA}_j=[\textbf{Q}; \textbf{A}_j]$}. Therefore the task aim is to find the most proper question-answer pair according to the context. 
	
\subsection{Extracting Knowledge}
    First, we extract the knowledge sources from an external knowledge graph, ConceptNet \cite{speer2017conceptnet}.\footnote{\textcolor{black}{We selected ConceptNet because it is the most widely-used corpus in the related studies, which is also well formed as the structural commonsense knowledge network that suits our task.}} Items with weight less than a threshold or contain words that are not in the vocabulary of the chosen PrLM are removed from KG. The items are triples with the form \{\textit{relation, head, tail}\}, which are rewritten as facts (\textit{e.g. \{causes, virus, disease\}} to \textit{virus causes disease}). These facts are encoded with our adopted PrLM and the last hidden states $\textbf{H}_k$ ($\textbf{H}_k\in \mathbb{R}^{n_k\times d_{model}}$ where $\textit{n}_k$ denotes the number of tokens in the fact) are taken as the output so that the representations of knowledge and context are in the same vector space. A self-attention module is used to refine the representation of each fact. We use mean-pooling in the end to aggregate the representation of each token and get a final representation $r_k$ ($r_k \in \mathbb{R}^{d_{model}}$) for each fact.
	
	\begin{equation}
	\begin{aligned}
	&\textrm{SelfAttention}(\textbf{H}_k)=\textrm{MHA}(\textbf{H}_k,\textbf{H}_k,\textbf{H}_k),\\
	&r_k=\textrm{mean}(\textrm{SelfAttention}(\textbf{H}_k)).
	\end{aligned}
	\end{equation}
	
	\subsection{Retrieve Relevant Knowledge}
	Each utterance $\textbf{U}_i$ is annotated with part-of-speech (POS) tags by NLTK \cite{loper2002nltk}.  For tokens with POS like adjectives, nouns, and verbs, we assume that they contain more implicit information than others; thus, items related to them are retrieved in KG. In all the chosen items, top $p$ (a hyperparameter) ones are selected to enhance the context representation. The extracted knowledge items are denoted as $\textbf{CK}=[r_{c_1},...r_{c_p}]$.
	
	For a QA-pair $\textbf{QA}_j$, we follow the same steps in dealing with $\textbf{U}_i$ to get the relevant knowledge items:
	\begin{equation}
	\textbf{QAK}_j=[r_{j_1},r_{j_2},...r_{j_k}], 
	\end{equation}
	where $j_k$ is the number of chosen knowledge items for $\textbf{QA}_j$.
	
	
	\subsection{Encoding and Representation Refinement}
	For each $\textbf{QA}_j$, it is concatenated with $\textbf{C}$ as input encoded with PrLM. The last hidden states $\textbf{H}_t$ are then separated into context representation $\textbf{H}^C$ and QA-pair representation $\textbf{H}^{QA}$. 
	
	Since we have questions in conversational reading comprehensions, which can serve as better indicators for selecting the pivot utterances. Therefore, the pivot selection is lightly different, which is based on the matching scores between each utterance with respect to the question instead. The representation of pivots turns $\textbf{H}^{P}$ is extracted from $\textbf{H}^C$ based on the position of key utterances as described in Section \ref{method:selection}. We use \textrm{MHA(·)} to calculate the pivot-refined context representation and  $\textbf{H}^{P}$. Simlilarly, we get the knowledge-refined representation of context and QA-pair:
	
	\begin{equation}
	\begin{aligned}
	&\textbf{H}^{CP}=\textrm{MHA}(\textbf{H}^{C},\textbf{H}^{P},\textbf{H}^{P}),\\
	&\textbf{H}^{CK}=\textrm{MHA}(\textbf{H}^C,\textbf{CK},\textbf{CK}),\\
	&\textbf{H}^{QA}=\textrm{MHA}(\textbf{H}^{QA},\textbf{QAK},\textbf{QAK}).
	\end{aligned}
	\end{equation}

	\subsection{Representation Fusion}
	Following \citet{zhu2020dual}, we use a Dual Multi-head Co-Attention (DUMA) module to fuse the representation of context and QA-pair.
	\begin{equation}
	\begin{aligned}
	&\textrm{MHA$_1$}=\textrm{MHA}(\mathbf{H}^C,\mathbf{H}^{QA},\mathbf{H}^{QA}),\\
	&\textrm{MHA$_2$}=\textrm{MHA}(\mathbf{H}^{QA},\mathbf{H}^{QA},\mathbf{H}^C),\\
	&\textrm{DUMA}(\mathbf{H}^C, \mathbf{H}^{QA}) = \textrm{Concat(mean(MHA$_1$)}\\&\textrm{,mean(MHA$_2$))}.
	\end{aligned}
	\end{equation}
	
	Based on different represenation of context and QA-pair calculated above, DUMA module may give three types of outputs, the original $\textbf{O}^O$, the pivot utterances refined $\textbf{O}^{P}$ and the knowledge refined $\textbf{O}^K$ as follow.
	\begin{equation}
	\begin{aligned}
	&\textbf{O}^O=\textrm{DUMA}(\textbf{H}^C,\textbf{H}^{QA}),\\
	&\textbf{O}^{P}=\textrm{DUMA}(\textbf{H}^{CP},\textbf{H}^{QA}),\\
	&\textbf{O}^K=\textrm{DUMA}(\textbf{H}^{CK},\textbf{H}_{QA}).\label{eq:know}
	\end{aligned}
	\end{equation}
	
	Then these three kinds of outputs are fused together as the final output. $\textbf{O}^P$ and $\textbf{O}^{K}$ are concatenated together and mapped to dimension of $2d_{model}$ through a linear layer to get the knowledge-pivot-utterances refined (KPR) output $\mathbf{O}^{KPR}$. Then we fuse the original output $\mathbf{O}^{O}$ and the KPR output $\mathbf{O}^{KPR}$ to get the final output $\mathbf{O}$. Concatenation is chosen as our fuse function.
	
	\subsection{Decoding}
	Our model decoder takes $\mathbf{O}$ and computes the probability distribution over answer options. Let $\mathbf{A}_i$ be the $i$-th answer option and $\mathbf{O}_i$ is the corresponding output of $<\mathbf{C},\mathbf{Q},\mathbf{A}_i>$. The loss function is computed by
	
	\begin{equation}
	L(\mathbf{A}_i|\mathbf{C},\mathbf{Q})=-\textrm{log}(\frac{\textrm{exp}(W^T\mathbf{O}_i)}{\sum_{j=1}^{l_a}\textrm{exp}(W^T\mathbf{O}_j)}),
	\end{equation}
	where $W$ is a learnable parameter.
	
\section{Experiment}\label{sec:exp}
\label{sec:length}
\subsection{Dataset}

We evaluated our model on three public multi-turn dialogue response selection datasets, the English Ubuntu Dialogue Corpus (Ubuntu) \cite{lowe2015ubuntu} and two Chinese datasets, namely the Douban Conversation Corpus (Douban) \cite{Wu2016Sequential} and E-commerce Dialogue Corpus (ECD) \cite{zhang2018modeling} to evaluate our backbone PoDS framework demonstrated in Section \ref{sec:dsm}, and one conversational comprehension dataset, i.e., DREAM \cite{sun2019dream} to assess our knowledge-enhanced variant described in Section \ref{model:knowledge}.\footnote{We also tried to employ the latter method for the three conventional response task; however, we did not see any performance gains. The reason is very likely that the three datasets are concerning technical discussion, social media, and e-commerce, which do not require much commonsense to solve the task.}

\noindent \textit{1) Dialogue Response Selection}

\paragraph{Ubuntu Dialogue Corpus} Ubuntu Dialogue Corpus comprises multi-turn human-computer conversations constructed from chat logs of the Ubuntu forum. It contains 1 million context-response pairs for training and 0.5 million pairs for validation and testing. The training set contains context-response pairs labeled as a positive or negative response randomly selected on the dataset. In both validation and test sets, each context contains one positive response and nine negative responses. 

\paragraph{Douban Conversation Corpus} Douban Conversation Corpus is an open-domain dataset constructed by the Douban Group, which provides a popular social networking service in China. Similarly constructed as the  Ubuntu corpus, this corpus contains 1 million context-response pairs for training, 0.5 million pairs for validation, and 6670 pairs for testing. 

\paragraph{E-commerce Dialogue Corpus} E-commerce Dialogue Corpus is a dataset of real-world conversations between customers and customer service staff. It contains 1 million context-response pairs for training and 10,000 pairs for both validation and testing. A topic has at least five types of conversation (e.g., commodity consultation, logistics discussions, recommendations, negotiations, and chat) relating to more than 20 commodities. The positive-to-negative ratio is 1:1 in training and validation and 1:9 in testing.\\

\noindent \textit{2) Conversational Reading Comprehension}

\paragraph[d]{DREAM}
DREAM is a newly released dialogue-based multi-choice MRC dataset, which is collected from English exams. Each dialogue, as the given context, has multiple questions, and each question has three response options. In total, it contains 6,444 dialogues and 10,197 questions. The most important feature of the dataset is that more than 80\% of the questions are non-extractive, and more than a third of the given questions involve commonsense knowledge. As a result, the dataset is small but quite challenging.

\subsection{Evaluation Metrics}
For the dialogue response selection tasks, we used the same evaluation metrics used in previous works \cite{Wu2016Sequential,zhang2018modeling}. 
Each model was evaluated  by  selecting  the $k$ best-matching responses from $n$ available candidates for the given conversation context. We calculate  the  recall  of  the  true  positive  replies  among  the $k$ selected  responses,  denoted $R_{}n@k$, as the main evaluation metric for each model. In addition, we used the mean average precision (MAP), mean reciprocal rank (MRR), and precision at position 1 (${P@1}$) for the Douban Conversation  Corpus. The  reason  for using the additional  metrics is that  the Douban  Conversation  Corpus  is  different  from  the  other three  datasets as it  includes  multiple  correct  candidates for a context in its test set, which may lead to low $R_{}n@k$.

For the conversational reading comprehension task, the evaluation metric we use is accuracy, \emph{acc=N$^+$/N}, where \textit{N$^+$} denotes the number of examples the model selects the correct answer, and \textit{N} denotes the total number of evaluation examples.

\begin{figure}
	\setlength{\abovecaptionskip}{0pt}
	\begin{center}
		\pgfplotsset{height=5.8cm,width=8.8cm,compat=1.14,every axis/.append style={thick},every axis legend/.append style={ at={(0.95,0.95)}},legend columns=1 row=2} \begin{tikzpicture} \tikzset{every node}=[font=\small] \begin{axis} [width=8.6cm,enlargelimits=0.13, xticklabels={1,2,3,4,5,6,7,8,9,10,11,12,13,14,15,16,17,18,19}, axis y line*=left, axis x line*=left, xtick={1,2,3,4,5,6,7,8,9,10,11,12,13,14,15,16,17,18,19}, x tick label style={rotate=0},
		ylabel={Percentage},
		ylabel style={align=left},xlabel={The last-$t$ utterance},font=\small]
		\addplot+ [smooth, mark=*,mark size=1.2pt,mark options={mark color=blue1}, color=blue1] coordinates { (1,0.088692) (2,0.238912) (3,0.138238) (4,0.094022) (5,0.075518) (6,0.062972) (7,0.047762) (8,0.036566) (9,0.028506) (10,0.025942) (11,0.023752) (12,0.025252) (13,0.026308) (14,0.02546) (15,0.02296) (16,0.017932) (17,0.012212) (18,0.006504) (19,0.00249)};
		\end{axis}
		\end{tikzpicture}
	\end{center}
	\caption{Proportion of the most related utterance (last-$t$) for gold response.}
	\label{prop}
\end{figure}
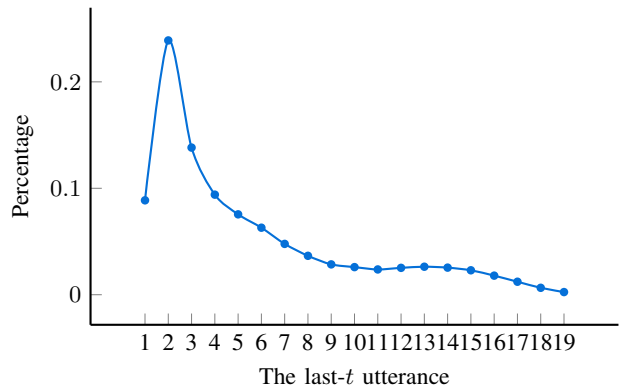

\subsection{Baseline Models}\label{baseline}
We used the pre-trained BERT as a baseline with its pairwise classification setting. In our study, we found it effective to use the context texts from the task-specific training set to fine-tune BERT with the language modeling objectives (Masked LM and Next Sentence Prediction) \cite{devlin2018bert} before training the PoDS, which we call domain fine-tuning (DFT). For DFT, the hyper-parameter setting is the same with the task training as demonstrated in Section \ref{sec:imp}.

We also compared our PoDS with the following published works.
\paragraph{Single-turn matching methods}
Single-turn matching models, including the RNN \cite{lowe2015ubuntu}, CNN \cite{lowe2015ubuntu}, LSTM \cite{lowe2015ubuntu}, BiLSTM \cite{kadlec2015improved}, MV-LSTM \cite{wan2016match}, and Match-LSTM \cite{wang2015learning}, concatenated all utterances in the context as a long document to calculate the matching score with a candidate response.

\begin{table*}\centering
	\caption{\label{tab:douban} Evaluation results of different models on the Douban Conversation Corpus and E-commerce Dialogue Corpus. 
	}
	{
		\begin{tabular}{l|cccc|cccccc|ccc}
			\hline
			\hline
			\multirow{2}{*}{\textbf{Model}}& \multicolumn{4}{c|}{\textbf{Ubuntu Dialogue Corpus}} & \multicolumn{6}{c|}{\textbf{Douban Conversation Corpus}} & \multicolumn{3}{c}{\textbf{E-commerce Dialogue Corpus}} \\
			 & ${\rm R_{2} @1} $  &${\rm R_{10} @1} $& ${\rm R_{10} @2} $ & ${\rm R_{10} @5} $ & MAP  &  MRR  & P@1 &${\rm R_{10} @1} $& ${\rm R_{10} @2} $ & ${\rm R_{10} @5} $&${\rm R_{10} @1} $& ${\rm R_{10} @2} $ & ${\rm R_{10} @5} $ \\
			\hline
			RNN &76.8 &40.3 &54.7 &81.9 &39.0 &42.2 &20.8 &11.8 &22.3 &58.9 &32.5 &46.3 &77.5\\
			CNN  & 84.8 &54.9 &68.4 &89.6 &41.7 &44.0 &22.6 &12.1 &25.2 &64.7 & 32.8 & 51.5 &79.2 \\
			LSTM & 90.1 &63.8 &78.4 &94.9  &48.5 &53.7 &32.0 &18.7 &34.3 &72.0 & 36.5 &53.6 &82.8\\
			BiLSTM & 89.5 &63.0 &78.0 &94.4 &47.9 &51.4 &31.3 &18.4 &33.0 &71.6 &35.5 &52.5 &82.5 \\
			DL2R & 89.9 &62.6 &78.3 &94.4 &48.8 &52.7 &33.0 &19.3 &34.2 &70.5 & 39.9 &57.1 & 84.2\\
			MV-LSTM  & 90.6 &65.3 &80.4 &94.6 &49.8 &53.8 &34.8 &20.2 &35.1 &71.0 &41.2 &59.1 &85.7  \\
			Match-LSTM & 90.4 &65.3 &79.9 &94.4  &50.0 &53.7 &34.5 &20.2 &34.8 &72.0 &41.0 &59.0&85.8 \\
			\hline
			Multi-View &90.8 &66.2 &80.1 &95.1 &50.5 &54.3 &34.2 &20.2 &35.0 &72.9&42.1 &60.1 &86.1 \\
			SMN & 92.6 &72.6 &84.7 &96.1 &52.9 &56.9 &39.7 &23.3 &39.6 &72.4 &45.3 &65.4 &88.6 \\
			DUA & - &75.2 &86.8 &96.2  &55.1&59.9&42.1&24.3&42.1&78.0 &50.1 &70.0 &92.1 \\
			DAM & 93.8 &76.7 &87.4 &96.9 &55.0&60.1&42.7&25.4&41.0&75.7 &52.6 &72.7 &93.3  \\
			IMN  & 94.6 &79.4 &88.9 &97.4 &57.0&61.5&44.3&26.2&45.2&78.9 &62.1 &79.7 &96.4\\
			MRFN & 94.5 &78.6 &88.6 &97.6 &57.1&61.7&44.8&27.6&43.5&78.3 &- &- &-\\
			IOI & 94.7 &79.6 &89.4 &97.4 &57.3&62.1&44.4&26.9&45.1&78.6 &56.3 &76.8 &95.0\\
			MSN & - &80.0 &89.9 &97.8 &58.7 &63.2&\textbf{47.0} &\textbf{29.5} &45.2 &78.8 &60.6 &77.0 &93.7\\
			\hline
			BERT  & 95.3 &81.7 &90.4 &97.7 &58.8 &63.1 & 45.3 &27.7 &46.4 &81.8 &62.1 &80.2 & 96.0 \\
			PoDS &  96.0 & 82.8  &  91.2 & 98.1  & 59.8  & 63.6  & 46.0  &  28.7 & 46.8  &  \textbf{84.5} &  63.3 & 81.0  &   96.7  \\
			+ DFT & \textbf{96.6} & \textbf{85.6} & \textbf{92.9} & \textbf{98.5} &  \textbf{59.9}  & \textbf{63.7}  & 46.0  &  28.7 &  \textbf{46.9} & 83.9 &  \textbf{67.1} & \textbf{84.2}  &   \textbf{97.3}  \\
			\hline
			\hline
		\end{tabular}
	}
	\smallskip
	\\ \footnotesize{Note: MSN \cite{yuan-etal-2019-multi} is the state-of-the-art model among published works. The best results are in boldface.}

\end{table*}

\paragraph{Multi-turn matching methods} Multi-turn matching models, including the multi-view model \cite{zhou2016multi}, DL2R \cite{yan2016learning}. SMN \cite{Wu2016Sequential}, DUA \cite{zhang2018modeling}, deep  attention  matching  network  (DAM) \cite{zhou2018multi}, IMN \cite{gu2019interactive}, MRFN \cite{tao2019multi}, IOI \cite{tao2019IOI}, and MSN \cite{yuan-etal-2019-multi}, matched the response with the utterances in the context.

\paragraph{Pre-trained Language Models}
Pre-trained language models, including BERT \cite{devlin2018bert}, XLNet \cite{yang2019xlnet}, RoBERTa \cite{liu2019roberta}, and ALBERT \cite{lan2019albert}.\footnote{Due to high computation cost, we only use widely-used BERT as the baseline for the three response selection tasks. For the conversational response selection task, we compare all these baselines results in Table \ref{tb:main}.}

\subsection{Implementation Details}\label{sec:imp}
Our implementations were based on the PyTorch version of BERT.\footnote{\url{https://github.com/huggingface/pytorch-pretrained-BERT}.} For the sake of training efficiency on the large corpora, we use \texttt{BERT-base-uncased} and \texttt{BERT-base-chinese} as initial weights on the English (Ubuntu) and Chinese datasets (Douban and ECD), respectively.\footnote{Since the corpora are quite large, training a BERT-based model requires a very long time, e.g., about 8 hours for one epoch on Ubuntu, though using the \texttt{BERT-base} models. Therefore, the SOTA results were reported on \texttt{base} models as well.} For the relatively smaller-scale DREAM dataset, we used ALBERT of both \texttt{base} and \texttt{xxlarge} variants \cite{lan2019albert} as our encoder, which is a recent dominant PrLM, to see if we can achieve even better performance though on such a strong PrLM model. \textcolor{black}{We set the $m$ to 12 by default, i.e., top-12 relevant utterances for context representation with careful consideration of effectiveness and efficiency (the analysis of $m$ will be presented in Section \ref{sec:num_pivots}.} We set the initial learning rate in \{1e-5, 2e-5, 3e-5\} with a warm-up rate of 0.1 and L2 weight decay of 0.01. The batch size was selected in \{24, 32, 64\}. The maximum number of epochs was set in [2, 5] depending on the dataset. Texts were tokenized using wordpieces, with a maximum length of 384 in all experiments. All our models were run on 32G NVIDIA V100 GPUs. We ran all the models up to 2 or 3 epochs and the best models on the dev set are chosen from all the checkpoints for test evaluation.

\begin{figure*}
	\centering
	\subfigure{
		\begin{minipage}[b]{0.3\linewidth}
			\setlength{\abovecaptionskip}{0pt}
			\begin{center}
				\pgfplotsset{height=5.5cm,width=6.5cm,compat=1.14,every axis/.append style={thick},every axis legend/.append style={ at={(0.95,0.95)}},legend columns=1 row=2} \begin{tikzpicture} \tikzset{every node}=[font=\small] \begin{axis} [width=6.5cm,enlargelimits=0.13, xticklabels={1,3,6,9,12,15,18,21}, axis y line*=left, axis x line*=left, xtick={1,2,3,4,5,6,7,8}, x tick label style={rotate=0},
				ylabel={${\rm R_{10} @1}$},
				ymin=85.18,ymax=85.9,
				ylabel style={align=left},xlabel={Top-$m$ selected utterances},font=\small]
				\addplot+ [smooth, mark=*,mark size=1.2pt,mark options={mark color=cyan}, color=red] coordinates { (1,85.544) (2,85.63) (3,85.648) (4,85.546) (5,85.55) (6,85.356) (7,85.304) (8,85.272)};
				\addlegendentry{\small PoDS}
				\addplot+[densely dotted, mark=none, color=cyan] coordinates {(1, 85.272)  (2, 85.272)  (3, 85.272)  (4, 85.272)  (5, 85.272)  (6, 85.272)  (7, 85.272)  (8, 85.272)};
				\addlegendentry{\small Baseline}
				
				\end{axis}
				\end{tikzpicture}
			\end{center}
			\centerline{(a) Accuracy on the Ubuntu dataset}
		\end{minipage}
	}
	\subfigure{
		\begin{minipage}[b]{0.3\linewidth}
			\setlength{\abovecaptionskip}{0pt}
			\begin{center}
				\pgfplotsset{height=5.5cm,width=6.5cm,compat=1.14,every axis/.append style={thick},every axis legend/.append style={ at={(0.95,0.95)}},legend columns=1 row=2} \begin{tikzpicture} \tikzset{every node}=[font=\small] \begin{axis} [width=6.5cm,enlargelimits=0.13, xticklabels={1,3,6,9,12,15,18,21}, axis y line*=left, axis x line*=left, xtick={1,2,3,4,5,6,7,8}, x tick label style={rotate=0},
				ylabel={MAP},
				ymin=59.18,ymax=60.2,
				ylabel style={align=left},xlabel={Top-$m$ selected utterances},font=\small]
				\addplot+ [smooth, mark=*,mark size=1.2pt,mark options={mark color=cyan}, color=red] coordinates { (1,59.69) (2,59.82) (3,59.63) (4,59.62) (5,059.25) (6,59.36) (7,59.32) (8,59.33)};
				\addlegendentry{\small PoDS}
				\addplot+[densely dotted, mark=none, color=cyan] coordinates {(1, 59.33)  (2, 59.33)  (3,59.33)  (4, 59.33)  (5, 59.33)  (6, 59.33)  (7, 59.33)  (8, 59.33)};
				\addlegendentry{\small Baseline}
				\end{axis}
				\end{tikzpicture}
			\end{center}
			\centerline{(b) Accuracy on the Douban dataset.}
		\end{minipage}
	}
	\subfigure{
		\begin{minipage}[b]{0.3\linewidth}
			\setlength{\abovecaptionskip}{0pt}
			\begin{center}
			\pgfplotsset{height=5.5cm,width=6.5cm,compat=1.14,every axis/.append style={thick},every axis legend/.append style={ at={(0.95,0.95)}},legend columns=1 row=2} \begin{tikzpicture} \tikzset{every node}=[font=\small] \begin{axis} [width=6.5cm,enlargelimits=0.13, xticklabels={1,3,6,9,12,15,18,21}, axis y line*=left, axis x line*=left, xtick={1,2,3,4,5,6,7,8}, x tick label style={rotate=0},
			ylabel={${\rm R_{10} @1}$},
			ymin=65.5,ymax=69.5,
			ylabel style={align=left},xlabel={Top-$m$ selected utterances},font=\small]
			\addplot+ [smooth, mark=*,mark size=1.2pt,mark options={mark color=cyan}, color=red] coordinates { (1,67.1) (2,66.4) (3,66.9) (4,67.4) (5,67.3) (6,65.5) (7,67.1) (8,66.0)};
			\addlegendentry{\small PoDS}
			\addplot+[densely dotted, mark=none, color=cyan] coordinates {(1, 66.0)  (2, 66.0)  (3,66.0)  (4, 66.0)  (5, 66.0)  (6, 66.0)  (7, 66.0)  (8, 66.0)};
			\addlegendentry{\small Baseline}
			\end{axis}
			\end{tikzpicture}
		\end{center}
		\centerline{(c) Accuracy on the ECD dataset.}
		\end{minipage}
	}
    \caption{\label{fig:select_number} Accuracy for a varying number of selected utterances.}
\end{figure*}
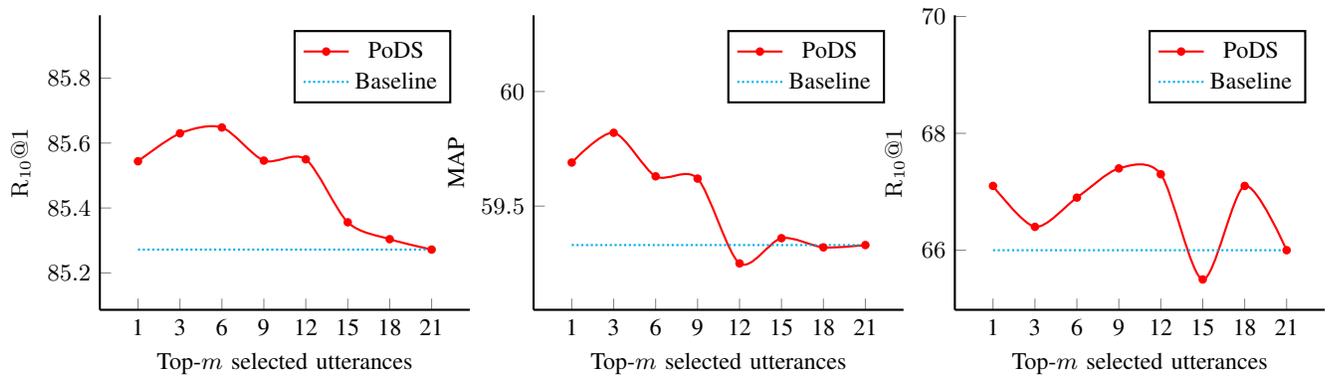
\subsection{Preliminary Experiments}\label{sec:plm}
Previous works \cite{zhang2018modeling,yuan-etal-2019-multi} heuristically selected the last utterance as the directive clue to refine the context, which showed substantial benefits to the multi-turn dialogue modeling. Intuitively, the last utterance would be instructive for the subsequent response. However, there are many complex multi-turn conversations that consist of jumping topics, as shown in Table \ref{tb-example-topic}. 

Inspired by the recent studies that enjoyed adopting pre-trained language models such as BERT to measure the semantic similarity between texts \cite{zhang2019bertscore,reimers-gurevych-2019-sentence}, we trained a multi-turn dialogue model using BERT to measure the average cosine similarity of each utterance and the gold response using the dev set of Ubuntu Dialogue Corpus \cite{lowe2015ubuntu}. Figure \ref{prop} shows the proportion of the most related (last-$t$) utterances for the gold response. We observe that, in most cases, the last three utterances are the most relevant to the intended response, which would be quite instructive for response selection. This observation verified the effectiveness of using the last utterance as the directive clue for context refinement and matching \cite{zhang2018modeling,yuan-etal-2019-multi} to some extent. However, we showed that the last one would not always be the best. This finding motivates us to investigate a more flexible way to select the most directive utterance(s) for fine-grained context modeling and context-response interactions. Besides the pivot utterances extracted from the given dialogue context, we are also interested in incorporating other indicators to improve the model capacity of conversation comprehension, such as external knowledge. 

\subsection{Main Results}
Table \ref{tab:douban} gives the evaluation results for the PoDS and baseline models on the traditional response selection tasks, showing that the PoDS outperformed the other models on all metrics and datasets. 
In particular, our model surpassed DAM \cite{zhou2018multi} by a large margin, where both models are based on a Transformer encoder. 
Moreover, our proposed model outperformed the strong BERT baseline substantially, achieving new state-of-the-art performance on all datasets. 

\textcolor{black}{We see that the recent models work relatively poorer in the Chinese datasets. The possible reason would be that those Chinese datasets are newer and more challenging datasets than the domain-specific Ubuntu dataset. Douban is an open domain conversation dataset that contains many more topics, and ECD is for the complex E-commence scenario that involves various types of conversations, e.g., commodity consultation, logistics express, recommendation, negotiation, and chitchat, over different commodities. Therefore, the Chinese tasks often involve more complex topic shifts and multiple intentions in a dialogue context, which require stronger models to solve the problems.}

\begin{table}
		
		\centering
		\caption{Results on DREAM dataset.
			\label{tb:main}
		}
		\setlength{\tabcolsep}{10pt}
		{
			\begin{tabular}{lll}
				\hline\hline
				\textbf{Model}                                                  & \textbf{Dev} & \textbf{Test} \\
				\hline
				FTLM++                                       & 58.1$^\star$     & 58.2$^\star$      \\
				BERT$_{large}$                             & 66.0$^\star$     & 66.8$^\star$      \\
				XLNet                                       & -         & 72.0$^\star$      \\
				RoBERTa$_{large}$                          & 85.4$^\star$     & 85.0$^\star$      \\
				RoBERTa$_{large}$+MMM                        & 88.0$^\star$     & 88.9$^\star$      \\
				ALBERT$_{xxlarge}$                        & 89.2$^\star$     & 88.5$^*$      \\
				ALBERT$_{xxlarge}$+DUMA                       & 89.3$^\dagger$     & \textbf{90.4}$^\dagger$      \\
				\hline
				ALBERT$_{base}$                                   & 67.4         & 67.3 \\
				ALBERT$_{base}$+\textcolor{black}{KPR}                                       & 69.3        & 68.7 \\
				ALBERT$_{xxlarge}$                               & 89.1          & 88.2  \\
				ALBERT$_{xxlarge}$+\textcolor{black}{KPR}                                   & \textbf{90.2}         & 89.8  \\

				\hline\hline        
			\end{tabular}%
		}
	\smallskip
	\\ \footnotesize{Note: Results denoted by $\star$ are from \citet{jin2019mmm}, $\dagger$ are from \citet{zhu2020dual}. \textcolor{black}{MMM is short for Multi-stage Multi-task Learning for Multi-choice Reading Comprehension \cite{jin2019mmm}}.}
	\end{table}
	
Table \ref{tb:main} gives the results on the DREAM dialogue comprehension dataset. Experimental results show our model obtains a great improvement compared to the baseline and achieves state-of-the-art performance for DREAM on dev set. 
	

\section{Analysis}\label{sec:ana}

\subsection{Effects of the Number of Selected Utterances}\label{sec:num_pivots}
Intuitively, the number of the pivot utterances $m$ would affect performance. We evaluated the performance of our model
for different numbers of selected sentences.
The comparison results are shown in Figures \ref{fig:select_number}. It is seen that the performance of the PoDS was remarkably improved when $m$ scales from 1 to 12. This indicates that the
information carried by one relevant utterance is commonly
insufficient to pinpoint the important part of the context and match with the response; thus, moderately selecting more utterances for matching may enhance the performance.
Meanwhile, this positive effect largely dissipated as $m$ increases from 12 to 20.
This is because an excessively large number of selected utterances may incorporate irrelevant or misleading information,
which hurt performance. Meanwhile, processing more
utterances would result in more computational cost. We thus selected the top-12 relevant utterances
for context representation with careful consideration of effectiveness and efficiency.\footnote{\textcolor{black}{We also considered setting a cosine similarity threshold and dividing the contexts into relevant and irrelevant parts. However, the scores we obtained are not quite distinguishable -- they either gathered around some specific scores, i.e., 0.9 or scattered irregularly. Therefore, we decided to use the empirical way of setting the ``hard threshold" with the fixed number of top-ranked utterances and achieved the performance gains.}}

\begin{table}
	\centering
	\caption{Comparison of different selection methods on the Ubuntu dataset. }
	\label{tab:selection_methods} 
	{
		\begin{tabular}{l|c|c|c}
			\hline
			\hline		
			\textbf{Model} & \textbf{${\rm R_{10} @1}$} & \textbf{${\rm R_{10} @2}$} & \textbf{${\rm R_{10} @5}$} \\
			\hline
			Baseline  &  85.2 & 92.5   &  98.2  \\
			PoDS &85.6  & 92.9  & 98.5 \\
			NLI &85.4  & 92.7  & 98.4 \\
			\hdashline
			CLS  &  85.5  & 92.8  & 98.6   \\
			Last  &  85.3 & 92.7   & 98.5   \\
			Random  &  85.3 & 92.6   & 98.3   \\
			POS &  85.5  & 92.7  &  98.5  \\
			\hline
			\hline
		\end{tabular}
	}

\end{table}
\subsection{Comparison of Different Selection Methods}\label{tab:selection_opt}
Besides the simple cosine similarity to measure the distance, Natural Language Inference (NLI) models also serve as an effective measure of semantic similarity \cite{khobragade2019machine}. We trained a BERT-based NLI model on the SNLI dataset \cite{bowman2015large} with 91.1\% dev accuracy, the linear layer has three output neurons for labels of \textit{contradiction}, \textit{entailment} and \textit{neutral}. We apply softmax on these outputs to get the probability of \textit{entailment} label as the similarity.

In addition to the similarity calculation algorithms, employing what kinds of patterns for selection would also matter, e.g., the special token used in BERT, \texttt{[CLS]}, which is supposed to carry the global information of the whole sequence, or with the last utterance. To investigate the influence of the selection method, we compared the results with different alternatives, 1) \textit{CLS}: used the representation of the special token \texttt{[CLS]} to replace the response representation for calculating the cosine similarity with each utterance; 2) \textit{Last}: directly used the last utterance as the pivot utterance; 3) \textit{Random}: randomly sampled an utterance as the pivot utterance; 4) \textit{Pos}: only calculated the cosine similarity for positive labeled responses during training. For the negative ones, we took the last utterance as the pivot. This aims to alleviate the noisy matching from the negative samples. Table \ref{tab:selection_methods} shows the results. \textcolor{black}{We see that the selection methods basically work better than the baseline, and it is possible to adopt alternatives for simplicity -- using \texttt{[CLS]} representation is a good alternative and only considering positive samples (POS) achieves a similar result.} Besides, using the last utterance shows to be suboptimal and random sampling is not a good practice.

\begin{table}
	\centering
		\caption{\label{tab:feed} Comparison of results obtained for different feeding patterns on the Ubuntu dataset. }
	{
		\begin{tabular}{l|c|c|c}
			\hline
			\hline		
			\textbf{Model} & ${\rm R_{10} @1} $& ${\rm R_{10} @2} $ & ${\rm R_{10} @5} $ \\
			\hline
			Baseline  &  85.2 & 92.5   &  98.2  \\
			Separate  & 84.9  & 92.3 & 98.1   \\
			Packed (PoDS) &85.6  & 92.9  & 98.5 \\
			\hline
			\hline
		\end{tabular}
		
	}

\end{table}

	\begin{figure}
		\centering
		\includegraphics[scale=0.68]{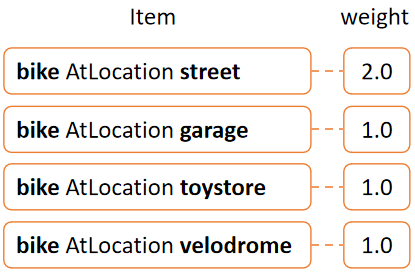}
		\caption{\textcolor{black}{Knowledge item in ConceptNet. Values on the right indicate the weight of each item in ConceptNet. A large weight means larger possibility.}}
		\label{fig:kt-concet}
	\end{figure}
	
\subsection{Discussion on the Matching Method}\label{match_method}
In this study, we found that the feeding pattern to the encoder affects the performance when it comes to PrLMs, like BERT. It is a natural idea to directly feed the separate utterance or response as individual input to the encoder, like in previous RNN-based response selection methods \cite{Wu2016Sequential,zhang2018modeling,zhou2018multi,tao2019IOI}. Then, the encoded representations are interacted by the same selection and matching mechanisms. Table \ref{tab:feed} shows the comparison of the packed and separate inputs. We found the benefit is trivial if we feed the separate inputs to BERT or use separate BERT embeddings to feed previous state-of-the-art models. The reason might be that the deep multi-head attention in BERT is effective for modeling the interactions of the paired input. We, therefore, recommend feeding the whole input that is packed with context and response and separating them later for further matching. On the basis of such a strong baseline, we also verified that we could yield further gains with our selective matching.

\subsection{Effects of External Knowledge Appending}
	
	Multi-turn dialogues more or less involve implicit or explicit commonsense when the conversation goes on; therefore, understanding them requires the support of knowledge. In our model, a knowledge graph is used by adding related knowledge items. In the example in Table  \ref{tb:example-commonsense} (in section \ref{sec:intro}), we need to know where a bike is likely to appear for answering the question. \textcolor{black}{As shown in Figure \ref{fig:kt-concet}}, the knowledge item \textit{\{atlocation, bike, street\}} can be found in our KG, which means \textit{Bikes are always found on the street}. It has a weight of 2, indicating it is more possible than others with a lower weight. With such a fact, the model thus can answer the question correctly.  
	
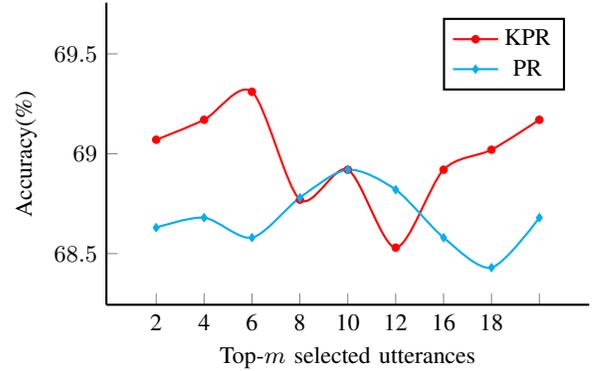
\begin{figure}
	\centering
			\setlength{\abovecaptionskip}{0pt}
			\begin{center}
			\pgfplotsset{height=5.6cm,width=6cm,compat=1.14,every axis/.append style={thick},every axis legend/.append style={ at={(0.95,0.95)}},legend columns=1 row=2} \begin{tikzpicture} \tikzset{every node}=[font=\small] \begin{axis} [width=8cm,enlargelimits=0.13, xticklabels={2,4,6,8,10,12,16,18}, axis y line*=left, axis x line*=left, xtick={1,2,3,4,5,6,7,8,9}, x tick label style={rotate=0},
			ylabel={Accuracy(\%)},
			ymin=68.4,ymax=69.6,
			ylabel style={align=left},xlabel={Top-$m$ selected utterances},font=\small]
			\addplot+ [smooth, mark=*,mark size=1.2pt,mark options={mark color=cyan}, color=red] coordinates
			{ (1,69.07) (2,69.17) (3,69.31) (4,68.77) (5,68.92) (6,68.53) (7,68.92) (8,69.02) (9,69.17)};
			\addlegendentry{\small KPR}
			\addplot+[smooth, mark=diamond*, mark size=1.2pt, mark options={mark color=cyan},  color=cyan] coordinates {(1, 68.63)  (2, 68.68)  (3,68.58)  (4, 68.78)  (5, 68.92)  (6, 68.82)  (7, 68.58)  (8, 68.43) (9,68.68)};
			\addlegendentry{\small PR}
			\end{axis}
			\end{tikzpicture}
		\end{center}
    	\caption{Comparison between complete model (KPR) and model without knowledge-refinement on different numbers of pivot utterances (PR).}
		\label{fig:com-kt}
\end{figure}

	To state the effects more clearly, we remove the knowledge refined output from our model \textcolor{black}{by dropping $\textbf{O}^K$ in Eq. \ref{eq:know}} and evaluate it on the dev set \textcolor{black}{of the DREAM dataset}. The results are shown in Figure \ref{fig:com-kt}. We can see a general performance improvement from the knowledge refined part for different numbers of pivot utterances.
	

	\begin{table}
		\centering
			\caption{Relevance score for each turn corresponding to the correct answer.}
		\begin{tabular}{ll}
			\hline\hline
			\textbf{Dialogue 1}                                                         & \textbf{Score} \\
			\hline
			\textit{W: Well, I'm afraid my cooking isn't to your taste.} & -1.91  \\ 
			\textit{\textbf{M: Actually, I like it very much.}}                         & \textbf{-1.49}                     \\
			\textit{W: I'm glad you say that. Let me serve you some more fish.}  & -2.53 \\
			
			\textit{\textbf{M: Thanks. I didn't know you were so good at cooking.}} & \textbf{-1.66}\\ 
			\textit{W: Why not bring your wife next time?}                              & -2.26                     \\
			\textit{M: OK, I will. She will be very glad to see  you, too.} & -1.87 \\
			\hline
			Question: \textit{What does the man think of the woman's cooking?}\\
			\hline
			\textit{A. It's really terrible.}                                                   \\
			\textit{B. It's very good indeed. *}                                        &                           \\
			\textit{C. It's better than what he does.}                                  &                          \\
			\hline\hline
		\end{tabular}%
		\label{tb:score}
	\end{table}

	\subsection{Effects of Pivots Utterance Extraction}
	
	As mentioned in the previous section, a challenge in understanding and modeling multi-turn dialogues is that the topic shifts in different turns, which means only a few turns are truly related to the question. 

	\textcolor{black}{Here, we select the NLI scores for better interpreting the benefits of utterance selection, because the numbers are more informative and distinctive for each utterance than the cosine similarity scores which either gather around some specific scores, i.e. 0.9, or scatter irregularly. Besides, the NLI selection method also achieves quite comparable results with the cosine one.}
	
	A higher relevance score indicates that it is more likely to conclude the QA given the corresponding turn. The example is shown in Table \ref{tb:score} verifies our hypothesis. Turns with top 2 entailment scores can directly give the answer, while the other turns have nothing to do with the answer. This example shows that key turns are decisive for context refinement in representation, and they are suggestive for explaining the contribution of each utterance.

	To address the effects more clearly, we evaluate our model by removing the key-turns refined output from our model and evaluate it. The maximum number of knowledge items is 30. The results are shown in Table \ref{tb:complete}, which indicates significant performance loss on both dev and test sets. The results verify that using the key turns as pivot utterances is indispensable for the advanced performance.

	
	
	
	\begin{figure}
	\centering
			\setlength{\abovecaptionskip}{0pt}
			\begin{center}
			\pgfplotsset{height=5.42cm,width=8cm,compat=1.14,every axis/.append style={thick},every axis legend/.append style={ at={(0.95,0.95)}},legend columns=1 row=2} \begin{tikzpicture} \tikzset{every node}=[font=\small] \begin{axis} [width=8cm,enlargelimits=0.13, xticklabels={2,4,6,8,10,12,16,18}, axis y line*=left, axis x line*=left, xtick={1,2,3,4,5,6,7,8,9}, x tick label style={rotate=0},
			ylabel={Accuracy(\%)},
			ymin=68.52,ymax=69.6,
			ylabel style={align=left},xlabel={Top-$m$ selected utterances},font=\small]
			\addplot+ [smooth, mark=*,mark size=1.2pt,mark options={mark color=cyan}, color=red] coordinates
			{ (1,69.07) (2,69.17) (3,69.31) (4,68.77) (5,68.92) (6,68.53) (7,68.92) (8,69.02) (9,69.17)};
			\addlegendentry{\small 30}
			\addplot+[smooth, mark=triangle*, mark size=1.2pt, mark options={mark color=cyan},  color=cyan] coordinates {(1, 69.07)  (2, 69.22)  (3,69.26)  (4, 68.97)  (5,  68.82)  (6, 68.58)  (7,  68.87)  (8, 69.01) (9,68.92 )};
			\addlegendentry{\small 60}
			\addplot+[smooth, mark=square*, mark size=1.2pt, mark options={mark color=orange}, color=orange] coordinates {(1, 69.07)  (2, 69.12)  (3,69.22)  (4, 68.97)  (5, 68.82)  (6, 68.63)  (7, 69.12)  (8,  69.02) (9,68.92)};
			\addlegendentry{\small 90}
			\end{axis}
			\end{tikzpicture}
		\end{center}
    	\caption{\textcolor{black}{Influence of number of knowledge items ($K$ items) on the dev set of DREAM dataset.}}
		\label{fig:kkt}
\end{figure}
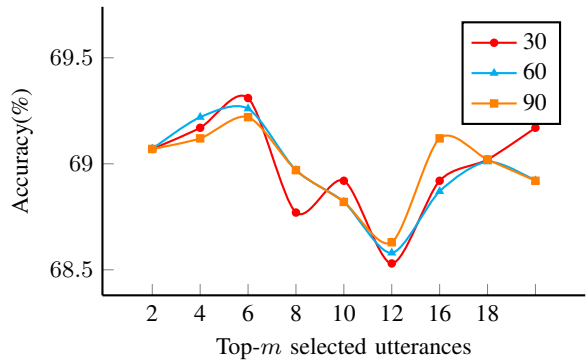

	
	\begin{table}
		\centering
		\caption{\label{tb:complete}Comparison between complete model, model without key-turns-refinement, and baseline.}
		\setlength{\tabcolsep}{6pt}
		{
			\begin{tabular}{lll}
				\hline\hline
				\textbf{Model}                     & \textbf{Dev} & \textbf{Test} \\
				\hline
				ALBERT$_{base}$                    & 67.40         & 67.31         \\
				\hdashline
			\textcolor{black}{ALBERT$_{complete}$(ALBERT$_{base}$+KPR)}                 & 69.32         & 68.71         \\
				-PR & 67.94        & 67.66       \\
				\hline\hline
			\end{tabular}%
		}
	\end{table}
	\subsection{Effects of Number of Knowledge Items}
	The number of knowledge Items can affect performance as well. So we evaluate our model on different numbers of knowledge items \textcolor{black}{on the dev set of the DREAM dataset}. The results are shown in Figure \ref{fig:kkt}. Contrary to our expectation, the results show little difference in various numbers of knowledge items. We suppose that the attention mechanism employed to refine context with external knowledge would contribute to the result because knowledge items with small weight tend to be concerned less in the attention mechanism as well since they are less relevant to the context. Therefore, the knowledge-refined context is similar though using more numbers of knowledge items, leading to a similar final performance when choosing a different number of knowledge items.  
	
	

\begin{table}
	\centering
	\makeatletter\def\@captype{table}\makeatother\caption{\textcolor{black}{An extracted example from the test set of Ubuntu where the baseline fails but is successfully solved by our model.}}
	{
		\begin{tabular}{p{6.8cm}}
			\hline 
			\textbf{Dialogue Context}\\
			\hline
			$\textbf{U}_1$: \textit{how do i do a real clean remove and install}                \\
			$\textbf{U}_2$: \textit{apt get/dpkg complain that its missing so i touch it}                         \\
			$\textbf{U}_3$: \textit{try to remove and start from scratch but the initscript is still missing}
			
			$\textbf{U}_4$: \textit{apt-get vs aptitude} \\ 
			
			$\textbf{U}_5$: \textit{thanks}                                    \\
			$\textbf{U}_6$: \textit{yeah that worked}                               \\
			\hline
	    	\textbf{PoDS:} 	\textit{thanks folks yeah but you re all so nice :-)}
		                     \\
			\textbf{Baseline:} \textit{maybe it cant be started from session init because of some particular status i really dont know :p ...}  \\
			\hline
		\end{tabular}%
	}
	
	\label{tb-error-ana}
\end{table}

\begin{figure}
	\centering
			\setlength{\abovecaptionskip}{0pt}
			\begin{center}
			\pgfplotsset{height=5.6cm,width=6cm,compat=1.14,every axis/.append style={thick},every axis legend/.append style={ at={(0.4,0.98)}},legend columns=1 row=2} \begin{tikzpicture} \tikzset{every node}=[font=\small] \begin{axis} [width=8cm,enlargelimits=0.13, xticklabels={2-4, 5-7,8-10}, axis y line*=left, axis x line*=left, xtick={1,2,3}, x tick label style={rotate=0},
			ylabel={${\rm R_{10} @1} $},
			ymin=0.8,ymax=0.88,
			ylabel style={align=left},xlabel={number of utterances},font=\small]
			\addplot+ [smooth, mark=*,mark size=1.2pt,mark options={mark color=cyan}, color=red] coordinates
			{ (1, 0.838382171)  (2, 0.849682233)  (3,0.864061864)};
			\addlegendentry{\small PoDS}
			\addplot+[smooth, mark=diamond*, mark size=1.2pt, mark options={mark color=cyan},  color=cyan] coordinates {(1, 0.800660338)  (2, 0.816155476)  (3,0.823990824) };
			\addlegendentry{\small Baseline}
			\end{axis}
			\end{tikzpicture}
		\end{center}
    	\caption{\textcolor{black}{${\rm R_{10} @1} $ of PoDS and the baseline BERT on different numbers of utterances.}}
		\label{fig:num_utterance}
\end{figure}
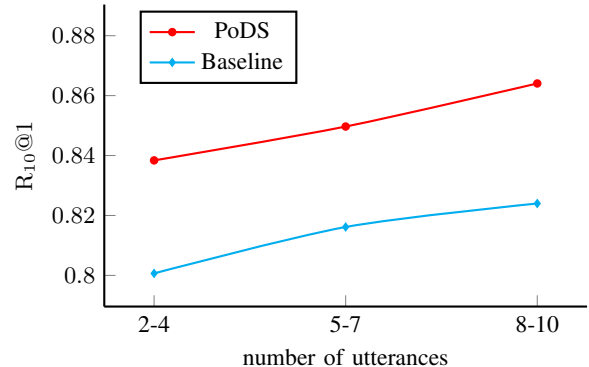

\textcolor{black}{\subsection{Prediction Analysis}}
\textcolor{black}{We analyzed the predictions from both of our system and the baseline on the test set of Ubuntu and DREAM datasets to understand how our model solves the error cases made by the baseline model, respectively.}

\textcolor{black}{Table \ref{tb-error-ana} shows an example from the Ubuntu test set. The response selected by PoDS is highly related to the topic flow of the utterances, from problem-solving to expressing the thanks, which indicates that our model is better at modeling the fluency of the dialogue, in other words, capturing the long-term relevance of the response and overall dialogue topic flow. For further exploration, we conduct an analysis by measuring the model performance on different context length that varies in different numbers of the utterances. Figure \ref{fig:num_utterance} shows that PoDS performs robustly and significantly than the baseline, especially for long contexts with more than 8 utterances.}

	\begin{figure*}
		\centering
		\includegraphics[width=1.0\textwidth]{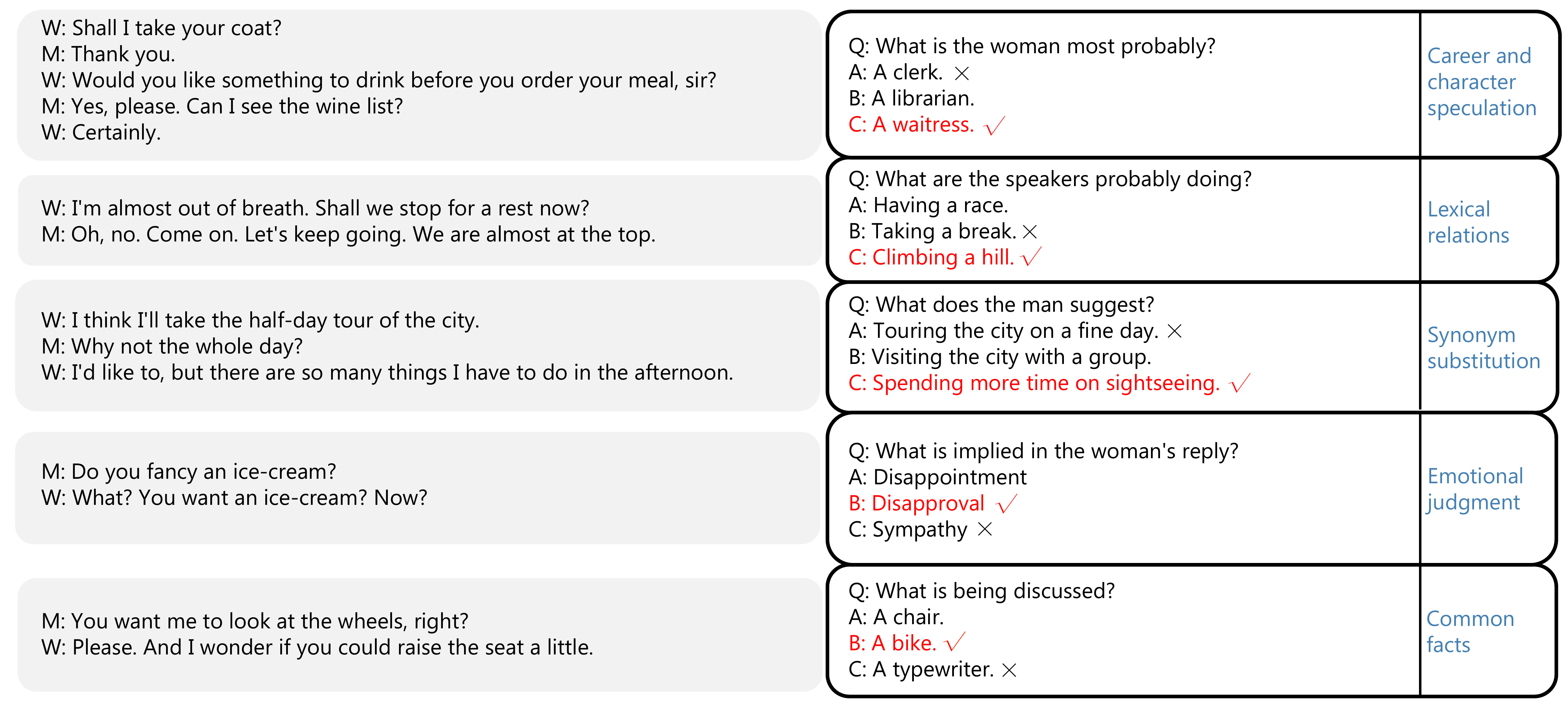}
		\caption{\textcolor{black}{Commonsense problems from the test set of DREAM where the baseline fails but is successfully solved by our model.}}
		\label{fig:ana_common}
	\end{figure*}

\textcolor{black}{For the more readable DREAM dataset that involves commonsense knowledge, we collected 92 examples that the baseline failed to answer while our model succeeded. }

\textcolor{black}{$\bullet$ 44.6\%: matching or summary cases which can be solved by simple text matching, extraction and searching.}

\textcolor{black}{$\bullet$ 25.0\%: logic consistency that involves complex reasoning.}

\textcolor{black}{$\bullet$ 6.5\%: arithmetic problems that involve mathematical calculation.}

\textcolor{black}{$\bullet$ 23.9\%: commonsense problems that include career and character speculation, synonymous substitution, lexical relations, emotional judgment based on modal particles, common facts, etc.}

\textcolor{black}{We observe that 23.9\% commonsense problems have been well solved by our model equipped with commonsense knowledge injection, as examples shown in Figure \ref{fig:ana_common}. Actually, the logic problems (25.0\%) also rely on commonsense, such as synonym substitution, antonym, and modal particle, to construct logic chains.}

\section{Conclusion}\label{sec:clu}
In this work, we proposed a pivot-oriented deep selection model using BERT as the encoder with pivot-aware contextualized attention mechanisms for the multi-turn response selection task. 
We first pick out some of the turns from the dialogue directly related to the candidate response or question as pivot utterances. Then, the relevant knowledge items are picked out and encoded with PrLM. The dialogue context is refined with pivot utterances and external knowledge items for better language representation, which is employed for the matching candidate response.
The procedures of our method are highly explainable and reflect a primary idea of cascading different models to get better language representation. 
Experimental results on four benchmark datasets show that our proposed model outperforms baseline models, achieving new state-of-the-art performance for multi-turn response selection. 
Case studies show that our selection strategies and the extra knowledge injection have  certain effectiveness for improving the model performance. 
In future work, we will investigate how to model the topic flow and logical consistency across multi-turn conversations to improve selection performance.


\ifCLASSOPTIONcaptionsoff
  \newpage
\fi



\bibliographystyle{plainnat}
%
\bibliography{reference}

%
\vspace{-12mm}
\begin{IEEEbiography}[{\includegraphics[width=1in,height=1.25in,clip,keepaspectratio]{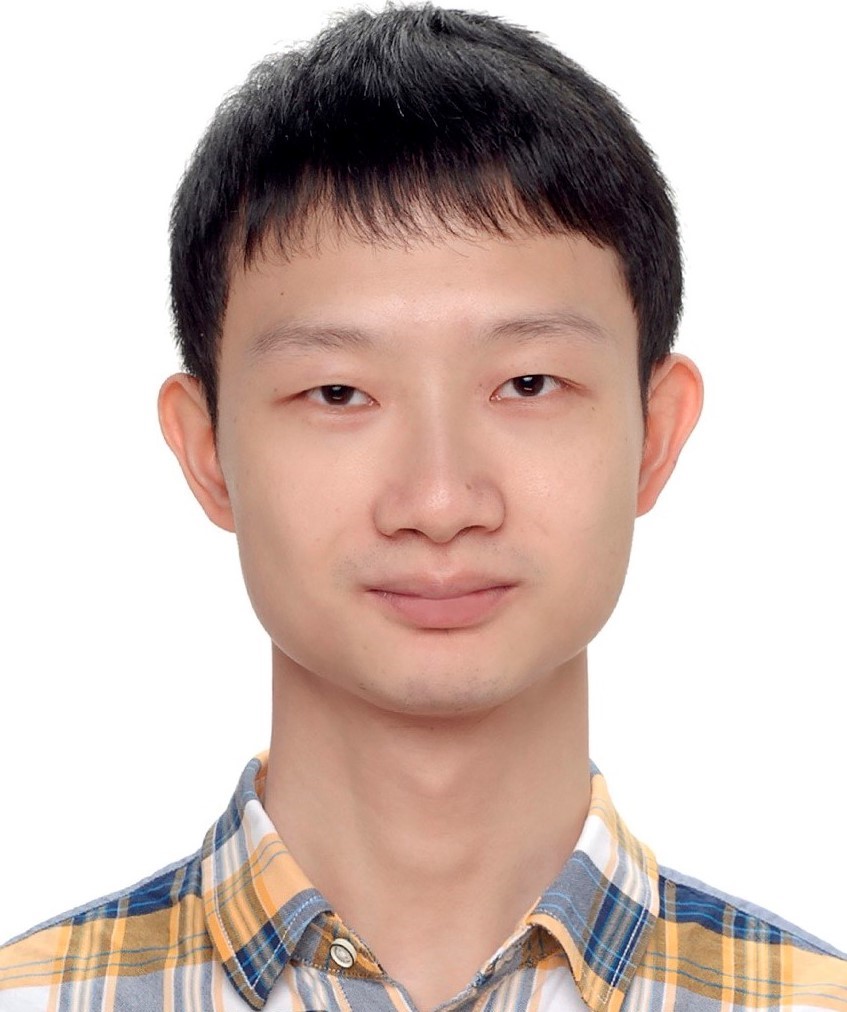}}]{Zhuosheng Zhang}
	received his Bachelor's degree in internet of things from Wuhan University in 2016, his M.S. degree in computer science from Shanghai Jiao Tong University in 2020. He is working towards his Ph.D. degree in computer science with the Center for Brain-like Computing and Machine Intelligence of Shanghai Jiao Tong University. He was an internship research fellow at NICT from 2019-2020. His research interests include natural language processing, machine reading comprehension, dialogue Systems, and machine translation. 
\end{IEEEbiography}
\vspace{-12mm}
\begin{IEEEbiography}[{\includegraphics[width=1in,height=1.25in,clip,keepaspectratio]{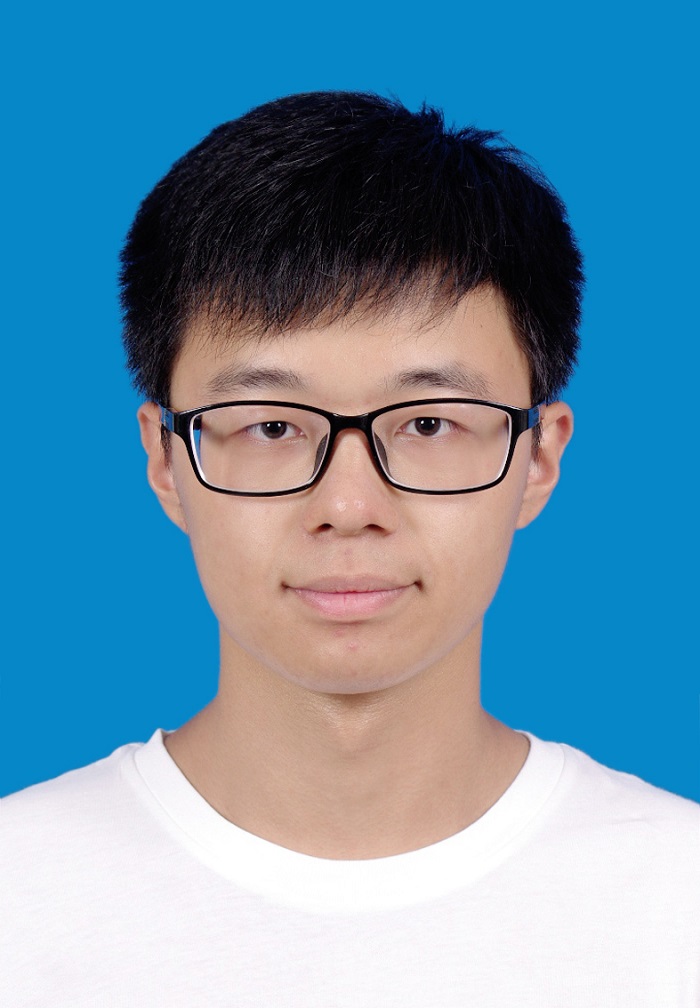}}]{Junlong Li} is an undergraduate student in the IEEE honor class of Shanghai Jiao Tong University. He majors in computer science and is expected to get his Bachelor's degree in 2022. His research interests include natural language processing, dialogue systems, and machine reading comprehension.
\end{IEEEbiography}
\vspace{-12mm}
\begin{IEEEbiography}[{\includegraphics[width=1in,height=1.25in,clip,keepaspectratio]{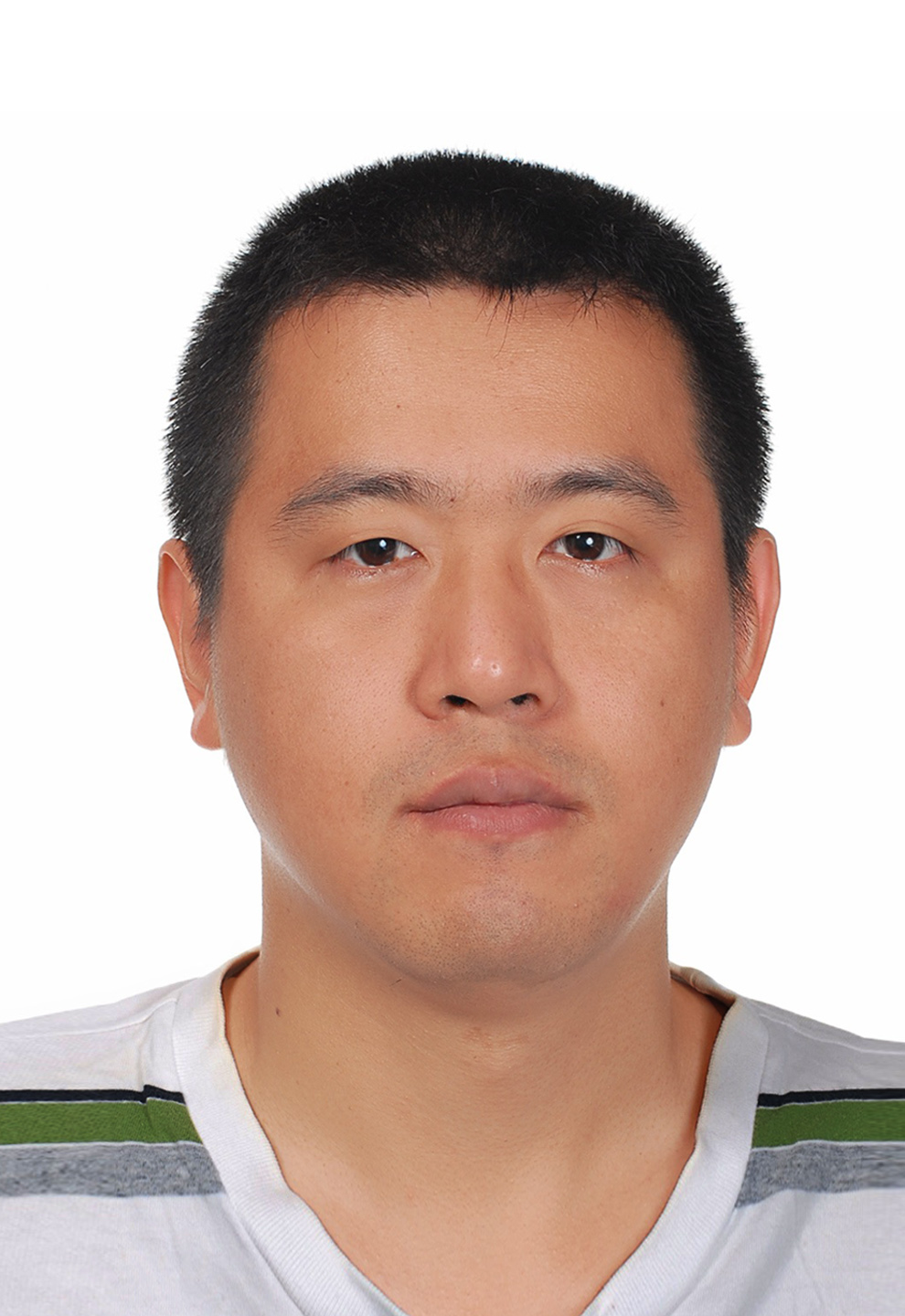}}]{Hai Zhao}
	received the BEng degree in sensor and instrument engineering, and the MPhil degree in control theory and engineering from Yanshan University in 1999 and 2000, respectively,
	and the PhD degree in computer science from Shanghai Jiao Tong University, China in 2005. 
	He is currently a full professor at department of computer science and engineering,  Shanghai Jiao Tong University after he joined the university in 2009. 
	He was a research fellow at the City University of Hong Kong from 2006 to 2009, a visiting scholar in Microsoft Research Asia in 2011, a visiting expert in NICT, Japan in 2012.
	He is an ACM professional member, and served as area co-chair in ACL 2017 on Tagging, Chunking, Syntax and Parsing, (senior) area chairs in ACL 2018, 2019 on Phonology, Morphology and Word Segmentation.
	His research interests include natural language processing and related machine learning, data mining and artificial intelligence.
\end{IEEEbiography}




\end{document}